\documentclass[11pt]{article} 
\usepackage{acl}
\usepackage{hyperref}
\usepackage{url}
\usepackage{verbatim}
\usepackage{multirow}
\usepackage{enumitem}  
\usepackage{latexsym}
\usepackage{booktabs}
\usepackage{makecell}

\usepackage{subfig}
\usepackage{microtype}
\usepackage[cmex10]{amsmath}
\usepackage{amsfonts, amsthm, amssymb}
\usepackage{pifont}
\newcommand{\cmark}{\textcolor{green}{\ding{51}}}%
\newcommand{\xmark}{\textcolor{red}{\ding{55}}}%
\usepackage{caption}
\usepackage{graphicx}  

\title{CMATH: Can Your Language Model Pass Chinese Elementary School Math Test?}
\author{Tianwen Wei\quad Jian Luan\quad Wei Liu\quad Shuang Dong \quad Bin Wang\\
Xiaomi AI Lab \\
\texttt{\{weitianwen,luanjian,liuwei40,dongshuang1,wangbin11\}@xiaomi.com}}

\begin{document}
\maketitle
\begin{abstract}

We present the Chinese Elementary School Math Word Problems (CMATH) dataset, comprising 1.7k elementary school-level math word problems with detailed annotations, source from actual Chinese workbooks and exams. This dataset aims to provide a benchmark tool for assessing the following question: to what grade level of elementary school math do the abilities of popular large language models (LLMs) correspond? We evaluate a variety of popular LLMs, including both commercial and open-source options, and discover that only GPT-4 achieves success (accuracy $\geq$ 60\%) across all six elementary school grades, while other models falter at different grade levels.
Furthermore, we assess the robustness of several top-performing LLMs by augmenting the original problems in the CMATH dataset with distracting information. Our findings reveal that GPT-4 is able to maintains robustness, while other model fail. We anticipate that our study will expose limitations in LLMs' arithmetic and reasoning capabilities, and promote their ongoing development and advancement.
\end{abstract}

\section{Introduction}
Recently, the field of artificial intelligence has witnessed groundbreaking advancements, particularly in the development of large language models (LLMs). Pioneering models such as ChatGPT \citep{instruct_gpt} along with \citep{galactica} have demonstrated impressing capabilities in understanding and generating natural language text across a multitude of tasks. The recently released GPT-4 \citep{gpt4_report, gpt4_sparks} model exhibits a sweeping range of skills, arguably far exceeding those of its predecessors and contemporaries. Its superior capabilities have unlocked new potential for application, not only in commercial settings but also in various scientific domains.

Mathematics, a core scientific discipline, represents a key area where the potential of LLMs can be harnessed. The ability to process, understand, and solve mathematical problems is a highly desirable trait for these models. This mathematical competence can lead to a myriad of applications, from providing assistance in educational contexts to facilitating complex computations in various sectors.

However, effectively evaluating the mathematical abilities of LLMs remains a non-trivial endeavor. Although several datasets have been developed for this purpose, they exhibit notable limitations. Firstly, most existing math-related datasets are in English \citep{gsm8k, mathqa, math}, making them unsuitable for evaluating Chinese language models. Secondly, many of these datasets present problems that are excessively difficult, e.g. college-level maths \citep{math, mmlu}, making them inappropriate for guiding the development of smaller language models. From our perspective, the most critical shortcoming is that the evaluation results derived from these datasets often lack intuitive clarity, making them challenging for the general public to comprehend. For instance, what does it truly mean when a model scores 35.7 on GSM8K \citep{gsm8k}? How can we interpret this score in terms of the model's mathematical competency?

We posit that the evaluation of LLMs should mirror that of human learners, which would allow us to convey results in a manner that is more intuitive and accessible. 
In pursuit of this human-centric evaluation, we introduce in this work the Chinese Elementary School Math Word Problems (CMATH) dataset, consisting of 1.7k elementary school-level math word problems sourced from actual Chinese workbooks and exams. 
Each problem in CMATH is annotated with grade information, enabling us to provide fine-grained evaluations akin to ``ChatGPT scored 70 out of 100 in a fourth-grade math exam''.

On our CMATH dataset, we conduct evaluation for a variety of popular LLMs, accessible via commercial API or released model weights. We discover that  GPT-4 is the only model that achieves success (accuracy $\geq$ 60\%) across all six elementary school grades.
We also examine the robustness of LLMs against the distracting information. It turns out that GPT-4 is again the sole model that maintains robustness, while other models are easily misled by the presence of distracting information.

\begin{figure*}[htbp]
\centering
\resizebox{0.85\textwidth}{!}{
\includegraphics{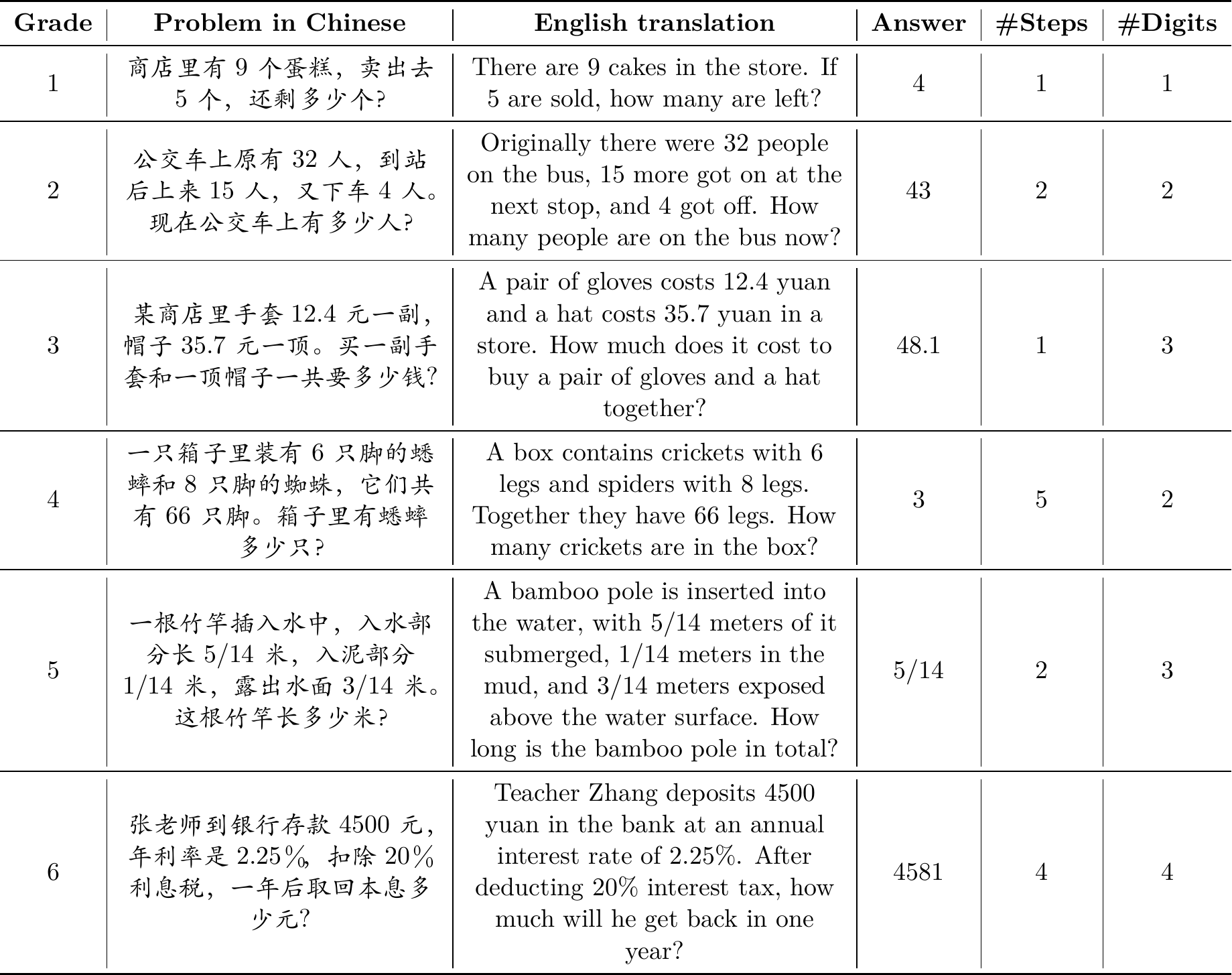}
}
\caption{Sample problems along with their English translations (not part of the dataset) and human annotations. The column title ``\#Steps'' and ``\#Digits'' stand for ``number of reasoning steps'' and ``number of digits'' respectively. }
\label{examples}
\end{figure*}

\section{CMATH dataset}
\subsection{Motivation}
This work is motivated by the following question:
\begin{quote}
To what grade level of elementary school math do the abilities of popular LLMs correspond?
\end{quote}
We create the CMATH dataset in order to  answer this question. We believe that the evaluation results of LLMs should be presented in an intuitive manner, making them easily understandable for the general public.

We are particularly interested in \emph{elementary school level} math word problems, as these problems, compared to high school or college level counterparts, provide a more appropriate evaluation of LLMs' \emph{general-purpose} reasoning and arithmetic capabilities. Elementary school math problems are more fundamental and, as a result, the skills required for solving them are more transferable to other domains. By assessing LLMs on these problems, we can gain valuable insights into their ability to generalize and adapt to new tasks.
Furthermore, the relatively simple nature of elementary school problems enhances their interpretability. It becomes easier to comprehend why an LLM succeeds or fails at solving these basic problems, allowing for a more transparent analysis of the underlying reasoning processes.

\subsection{Data Collection \label{dataset}}
We collect the math word problems from Chinese elementary school exercise books and exams that are freely available on the internet. 
The original data comes in as either PDF or Microsoft Word format, which is subsequently converted, preferably automatically, otherwise manually by human annotators into pure text. As we are only interested in text-based math word problems, we discard all problems originally equipped with graphic content. All questions also go through the standard data preprocessing pipeline, such as deduplication and cleaning. Following this, the questions undergo several rounds of human validation by the authors.

\begin{table}[t]  
\centering  
\begin{tabular}{c|cccc}  
\toprule
grade & size & length  & steps & digits  \\  \midrule
1    & 254  & 33.6 & 1.3 & 1.9 \\ 
2    & 353  & 35.5 & 1.6 & 2.0   \\ 
3    & 337  & 42.1 & 1.9 & 2.8  \\  
4    & 220  & 47.0 & 2.1 & 3.3   \\ 
5    & 227  & 48.9 & 2.7 & 2.7   \\  
6    & 298  & 52.5 & 3.0 & 3.2  \\ 
\bottomrule
\end{tabular}  
\caption{Statistics of the CMATH dataset.
The column titled ``length''  denotes the average problem length in terms of the number of characters.
The column titled ``steps'' denotes  the average reasoning steps required to solve the problem.
The column titled ``digits'' stands for  the average number of digits involved in the problem solution.}  
\label{stats}  
\end{table}  

\subsection{Data annotation}
We provide annotations for the collected problems, including grade, answer, number of reasoning steps and number of digits.
Examples can be found in Table \ref{examples}.
\subsubsection{Grade}
We annotate the elementary school grade to which each collected math word problem belongs. This information can be used to create subsets of problems specific to a particular grade, enabling more targeted, fine-grained evaluation results.

\subsubsection{Answer}
We annotate the ground truth answer for each problem. Annotated answers are standalone numerals that fall into one of the following categories: integer, decimal number, fraction, or percentage. We do not provide the reasoning process leading to the answer, as our dataset is intended for test-only purposes. 

\subsubsection{Number of Reasoning Steps}
For each problem, we manually annotate the number of reasoning steps required to solve it. 
This quantity is straightforward for the majority of problems, where human annotators can easily reach consensus (e.g., examples in Table \ref{examples} except for the one from grade 4). We acknowledge that, in a few cases, the number of steps may vary depending on the specific solution one considers (as with the problem of grade 4 in Table \ref{examples}). However, this ambiguity should not pose a serious issue, as it only accounts for a small fraction of problems. We use the number of reasoning steps as a suitable proxy for a problem's reasoning complexity, which relates to the level of logical analysis and problem-solving strategies needed for an LLM to arrive at the correct solution. Generally, more reasoning steps correspond to a more intricate thought process and potentially more opportunities for an LLM to make errors or lose track of the problem's structure.

\subsubsection{Number of Digits}
Each math word problem is associated with several numbers in the problem statement. For a given problem $P$, we denote the set of associated numbers by $\mathcal{N}$. LLMs are expected to perform a number of arithmetic operations on $\mathcal{N}$ to derive the final numerical answer $a$.

As a rough measure of the arithmetic complexity of $P$, we consider
\begin{equation}
D = \max\{\texttt{len}(x),\,\,\forall x\in\mathcal{N}\cup\{a\}\},
\end{equation}
where \texttt{len(x)} returns the number of digits\footnote{Only digits $0\sim9$ are counted. Other symbols, such as decimal points, percentage symbols, and slashes, are not taken into account.} in the string representation of $x$. In the following sections, we simply refer to $D$ as the \emph{number of digits} of $P$. This quantity is a practical and easily quantifiable measure of the computational demands placed on an LLM when tackling a problem.

We developed a simple program to automatically compute $D$ for each problem.

\section{Experimental Setup}

\begin{table}[h]
\centering
\begin{tabular}{lcc}
\toprule
\textbf{Model}  & \textbf{Parameters} & \textbf{Access} \\ 
\midrule
GPT-4  &  - & API \\   
ChatGPT &  - & API \\ 
\midrule
Chinese-Alpaca &  33B/13B & Weights    \\
Moss      & 16B  & Weights \\ 
Ziya-LLaMA-13B	   & 13B  & Weights        \\
Baichuan-7B    & 7B    & Weights \\
RWKV-7B        & 7B    & Weights \\
ChatGLM-6B     & 6B    & Weights \\ 
\bottomrule
\end{tabular}
\caption{Language models evaluated in this work. 
}
\label{models}
\end{table}

\subsection{Models}
We consider a variety of popular LLMs that are able to process text in Chinese and are fine-tuned to be general-purpose task solver. Those LLMs, being developed by diverse organizations, vary in size and can be accessed via either API or model weights as summarized in Table \ref{models}. 

\begin{itemize}[leftmargin=*]  
\item GPT-4 \citep{gpt4_report} is OpenAI's newest generation of LLM. It is arguably the most powerful LLM as of the time of writing this manuscript (June 2023) and is considered as the first artificial general intelligence (AGI \citet{gpt4_sparks}). However, the technical details are not disclosed.

\item ChatGPT is the predecessor of GPT4. It is based on InstructGPT \citep{instruct_gpt}, which has undergone instruction tuning and reinforcement learning from human feedback. The version of ChatGPT evaluated in this work is identified as ``{\tt gpt-3.5-turbo}'' in OpenAI's API.   The technical details of this model are not disclosed.




\item MOSS \citep{moss} is an open source LLM with 16B parameters based on CodeGen \citep{codegen}. It is further pre-trained on 100B Chinese tokens and 20B English Tokens, then fine-tuned on 110M multi-turn conversational data. 

\item Ziya-LLaMA-13B \citep{ziya} is based on LLaMA-13B, where the original vocabulary is extended with 7k Chinese characters, and the checkpoint is further pre-trained on 110B tokens of Chinese text. 
After the continual pre-training, Ziya-LLaMA-13B has also undergone RLHF training as in \citep{instruct_gpt}.

\item Chinese-Alpaca \citep{chinese-alpaca} is also based on LLaMA series, extended with Chinese vocabulary. The model is has undergone supervised instruction tuning on Chinese datasets with Low Rank Adaptation (LoRA) technique \citep{lora}. In this work we evaluate 13B and 33B versions of Chinese-Alpaca.

\item RWKV-7B \citep{RWKV} is an RNN-Transformer hybrid model with 7B parameters. The model is pre-trained on both English and Chinese texts, and is fine-tuned on open source instruction-tuning datasets.  More information can be found in \citep{rwkv-huggingface}. 

\item Baichuan-7B \citep{baichuan} is a LLaMA-like LLM pre-trained from scratch on 1.2T Chinese and English tokens. Although it is merely a foundation model, in preliminary experiments we find out that it is able to solve math word problems in a zero-shot manner.  Therefore we also evaluate its performance in this work.

\item ChatGLM-6B  \citep{chatglm6b} and its successor ChatGLM2-6B \citep{chatglm6b-v2} feature a modified encoder-decoder transformer architecture \citep{glm, glm130}. The two models are pre-trained on English and Chinese data and undergoes supervised instruction tuning. 

\end{itemize}

\subsection{Evaluation Procedure}
\subsubsection{Zero-shot Evaluation}
Throughout our experiments we employ the zero-shot evaluation method, eschewing any form of supplementary prompting. This entails presenting the problem statement in its original form to the LLM to obtain the model response. 
We deliberately forgo prompting-based evaluation approaches, such as few-shot or chain-of-thought (CoT, \citet{cot}) evaluations, as the LLMs examined in this study are all fine-tuned models intended for direct deployment in real-world applications.
We posit that zero-shot evaluation furnishes a more accurate and pragmatic assessment of model performance.

\subsubsection{Automated Evaluation}
Given an input of math word problem, a typical LLM generated response encompasses several paragraphs detailing reasoning steps culminating in a final answer. To ascertain the correctness of the model's answer, we employ a regular expression-based script to extract all numerals within the response that are likely to constitute the concluded answer. We then compare these extracted numerals with the annotated ground truth answer, deeming the LLM's solution correct if a numerical match is identified between the ground truth and any of the extracted figures.

We have scrutinized the accuracy of our automated evaluation procedure by manually examining a random sample of 200 problems and LLM-generated responses. Our findings reveal that the precision and recall of our method stand at 96\% and 98\%, respectively.

\begin{figure*}[htbp]
\centering
\subfloat[]{\label{fig:a}\includegraphics[width=0.4\textwidth]{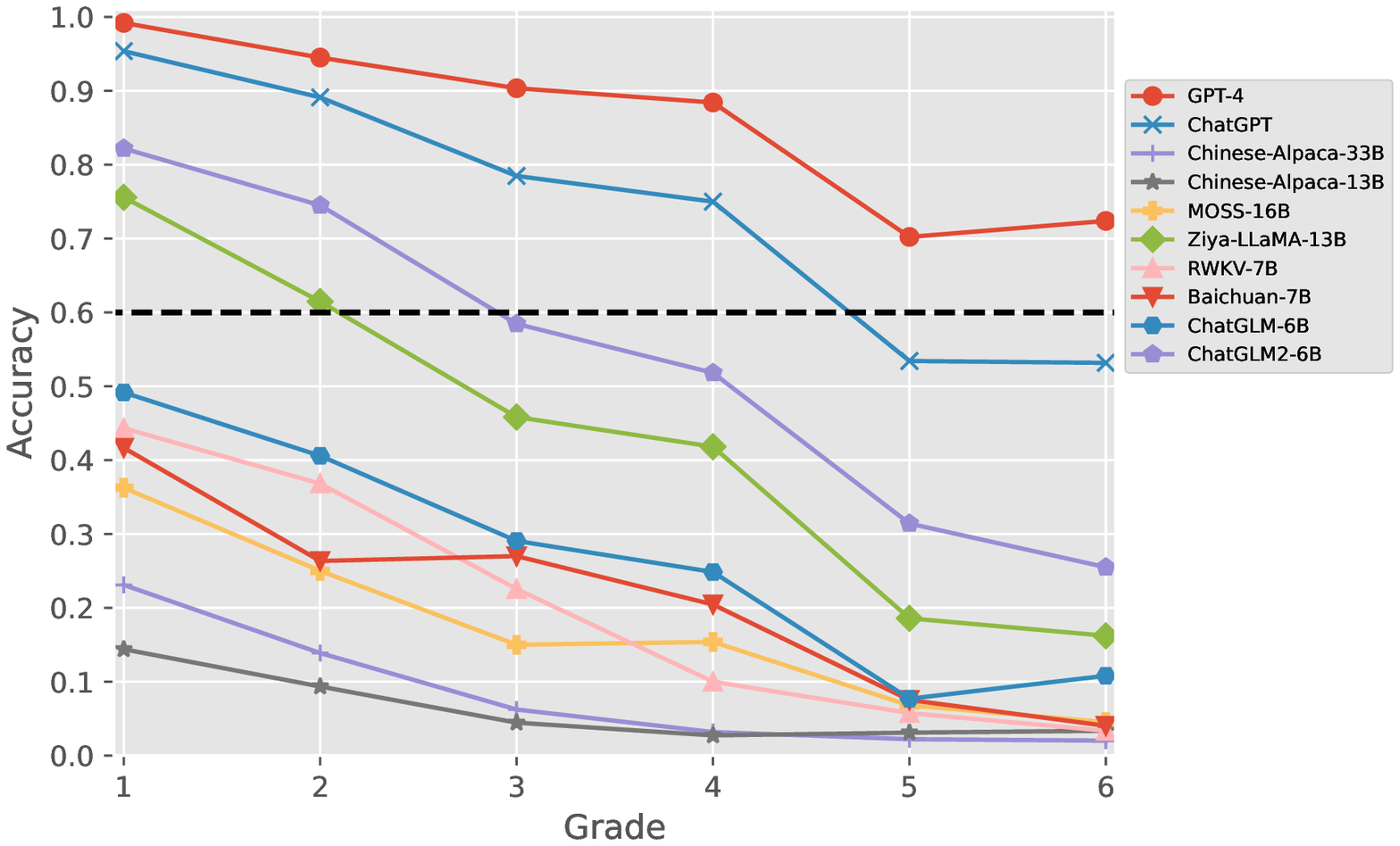}}
\subfloat[]{\label{fig:b}\includegraphics[width=0.4\textwidth]{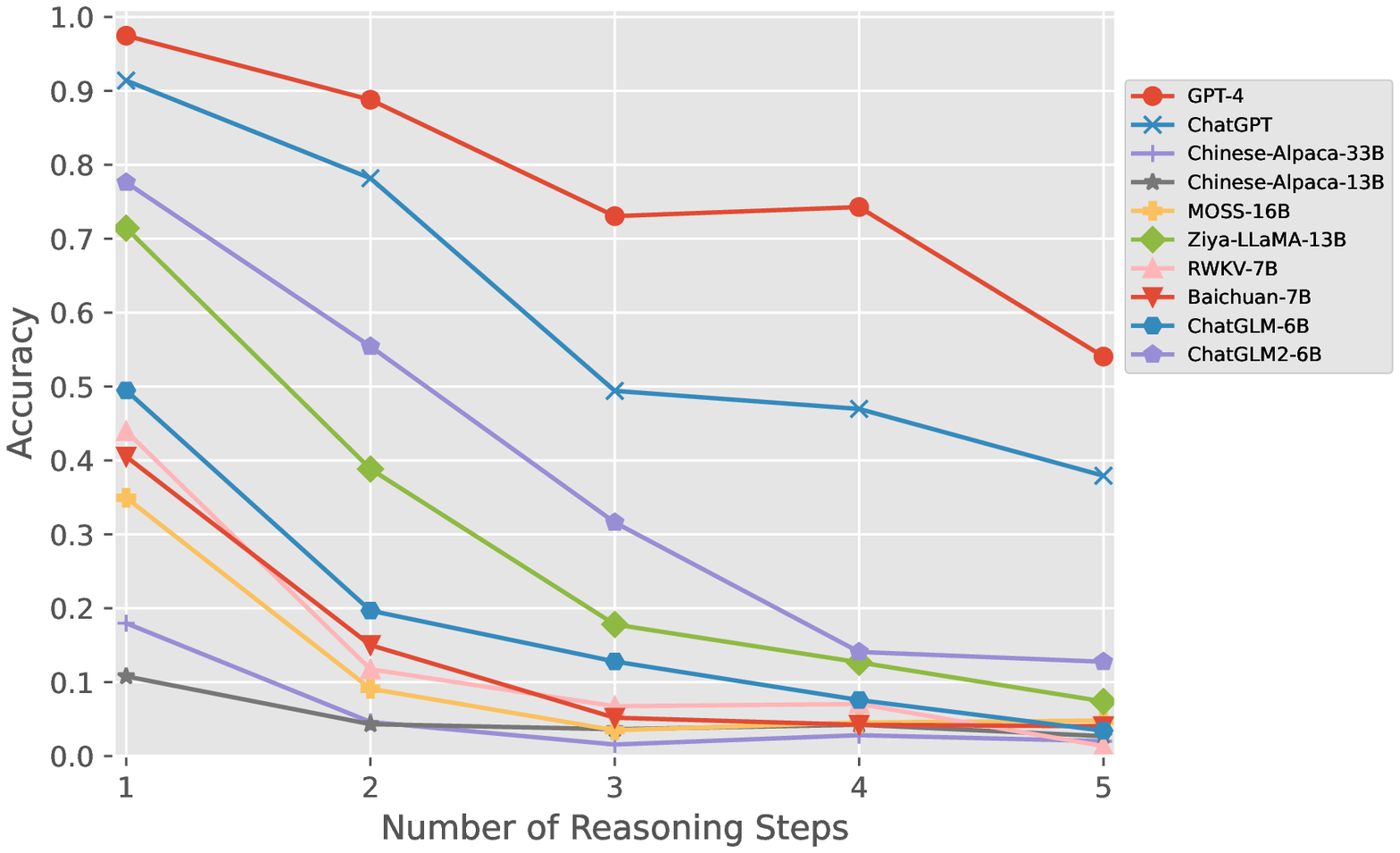}}\\
\subfloat[]{\label{fig:c}\includegraphics[width=0.4\textwidth]{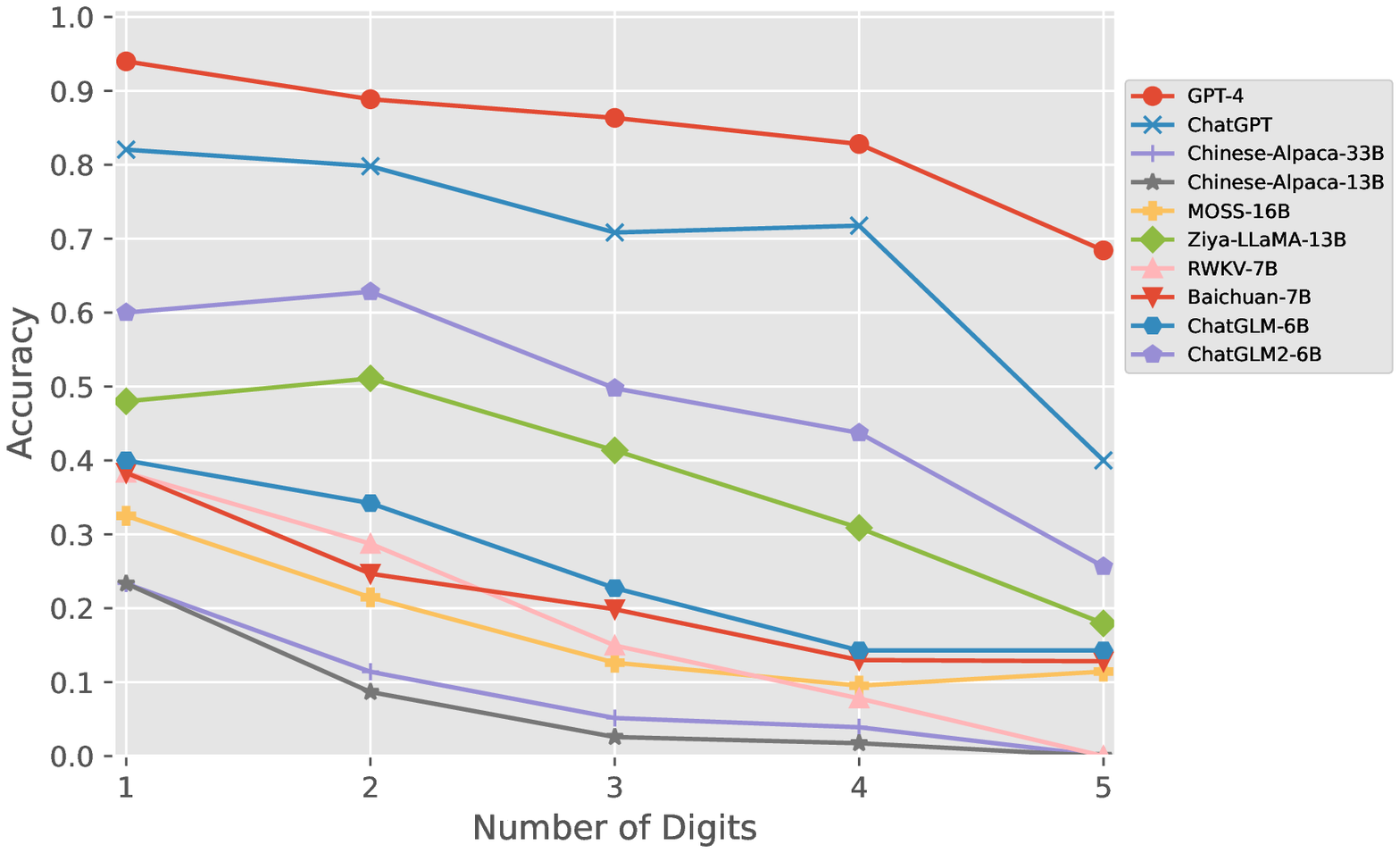}}
\subfloat[]{\label{fig:d}\includegraphics[width=0.4\textwidth]{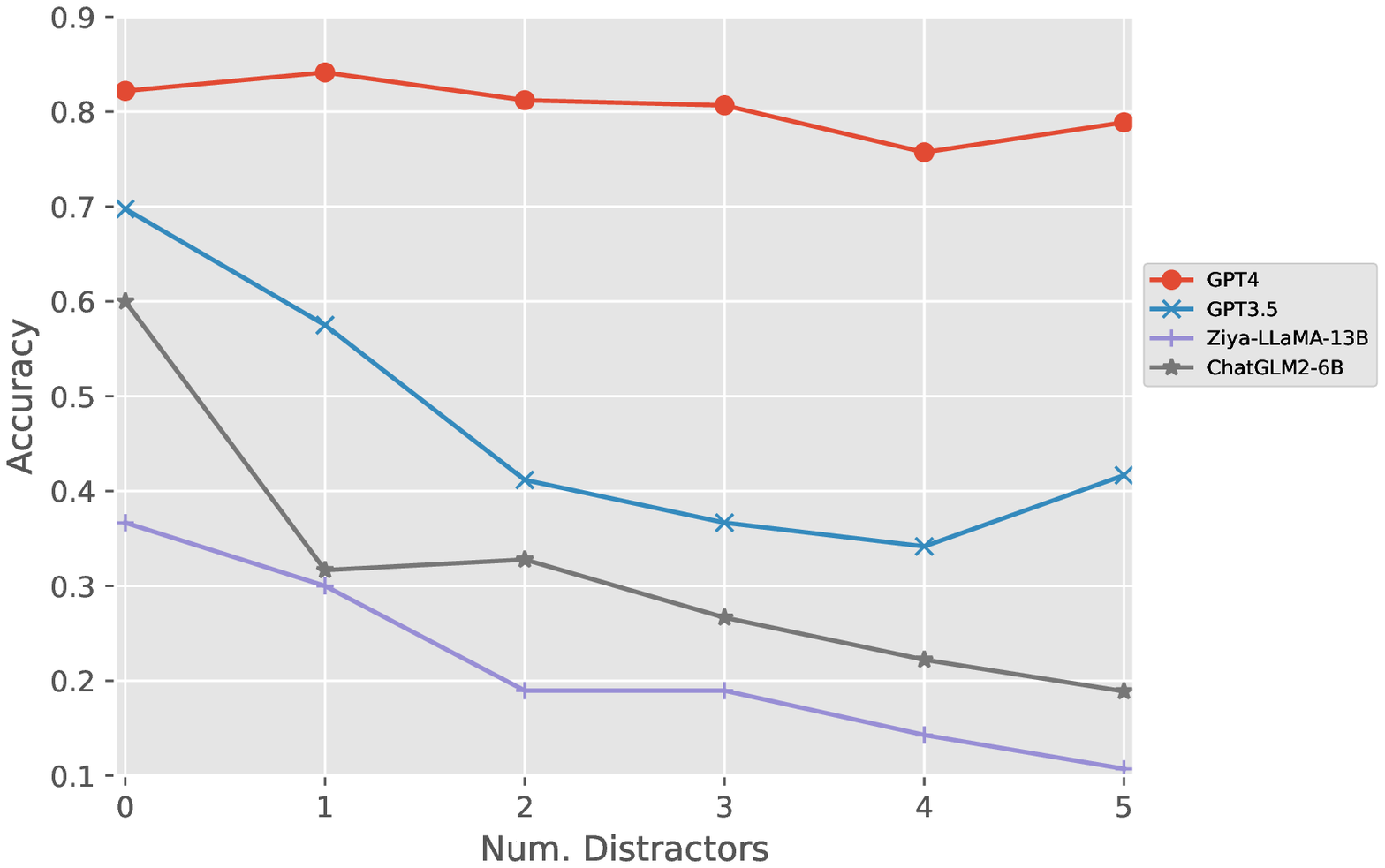}}
\caption{(a) (b) (c): The plot of average test accuracy against one of the problem complexity measures, including grade, number of reasoning steps and number of digits, for each LLM. (d): The plot of average test accuracy against the number of distractors on the distractor dataset, for the top performant models. }
\label{fig:main}
\end{figure*}

\section{Result and Analysis}
\subsection{Main results}
The test results\footnote{Results from API are obtained early June, 2023.} for are presented in Figure \ref{fig:main} (a), illustrating the accuracy per grade for each model.
 From the figure, a distinct downward trend in accuracy is evident, signifying that the performance of all models declines as the grade level increases. Although this outcome is somewhat anticipated, given that higher-grade math problems generally present greater difficulty,  it is still surprising to observe that half of the models struggle even at grade 1. 

GPT-4 emerges as the sole model capable of achieving success (accuracy exceeding 60\%) in math tests across all six elementary school grades. Following GPT-4, ChatGPT demonstrates success in tests for grades 1 to 4, but encounter difficulties in grades 5 and 6. The subsequent high-performing model is ChatGLM2-6B, succeeds only in grades 1 and 2 but displays impressive performance considering its size. The remaining models fail across all grade levels. 

Our results reveal that, despite being deemed relatively simple for an average human adult, math word problems at the elementary school level continue to pose challenges for general-purpose open source LLMs.

\begin{table}[h]  
\centering 
\resizebox{0.45\textwidth}{!}{
\begin{tabular}{l|llllll}  
\toprule
Model    & G1 & G2 & G3 & G4 & G5 & G6 \\ \midrule
GPT-4     & \cmark  & \cmark   & \cmark  & \cmark  & \cmark  & \cmark  \\ 
ChatGPT  & \cmark  & \cmark  & \cmark  & \cmark  & \xmark & \xmark \\ 
\midrule
Chinese-Alpaca-33B   & \xmark & \xmark & \xmark & \xmark & \xmark & \xmark \\ 
Chinese-Alpaca-13B   & \xmark & \xmark & \xmark & \xmark & \xmark & \xmark \\ 
MOSS-16B     & \xmark & \xmark & \xmark & \xmark & \xmark & \xmark \\ 
Ziya-LLaMA-13B     & \cmark & \cmark & \xmark & \xmark & \xmark & \xmark \\ 
RKWV-7B   & \xmark & \xmark & \xmark & \xmark & \xmark & \xmark \\ 
Baichuan-7B   & \xmark & \xmark & \xmark & \xmark & \xmark & \xmark \\ 
ChatGLM-6B   & \xmark & \xmark & \xmark & \xmark & \xmark & \xmark \\ 
ChatGLM2-6B   & \cmark & \cmark & \xmark & \xmark & \xmark & \xmark \\ 
\bottomrule
\end{tabular}  
}
\label{results}  
\caption{
Results indicating whether LLMs succeed or fail in solving math problems from each grade level. In the table, G1 to G6 denote grade levels 1 to 6, while \cmark \, and \xmark \, represent success and failure, respectively.
} 
\end{table}

\begin{table*}[htbp]
\centering
\resizebox{0.9\textwidth}{!}{
\includegraphics{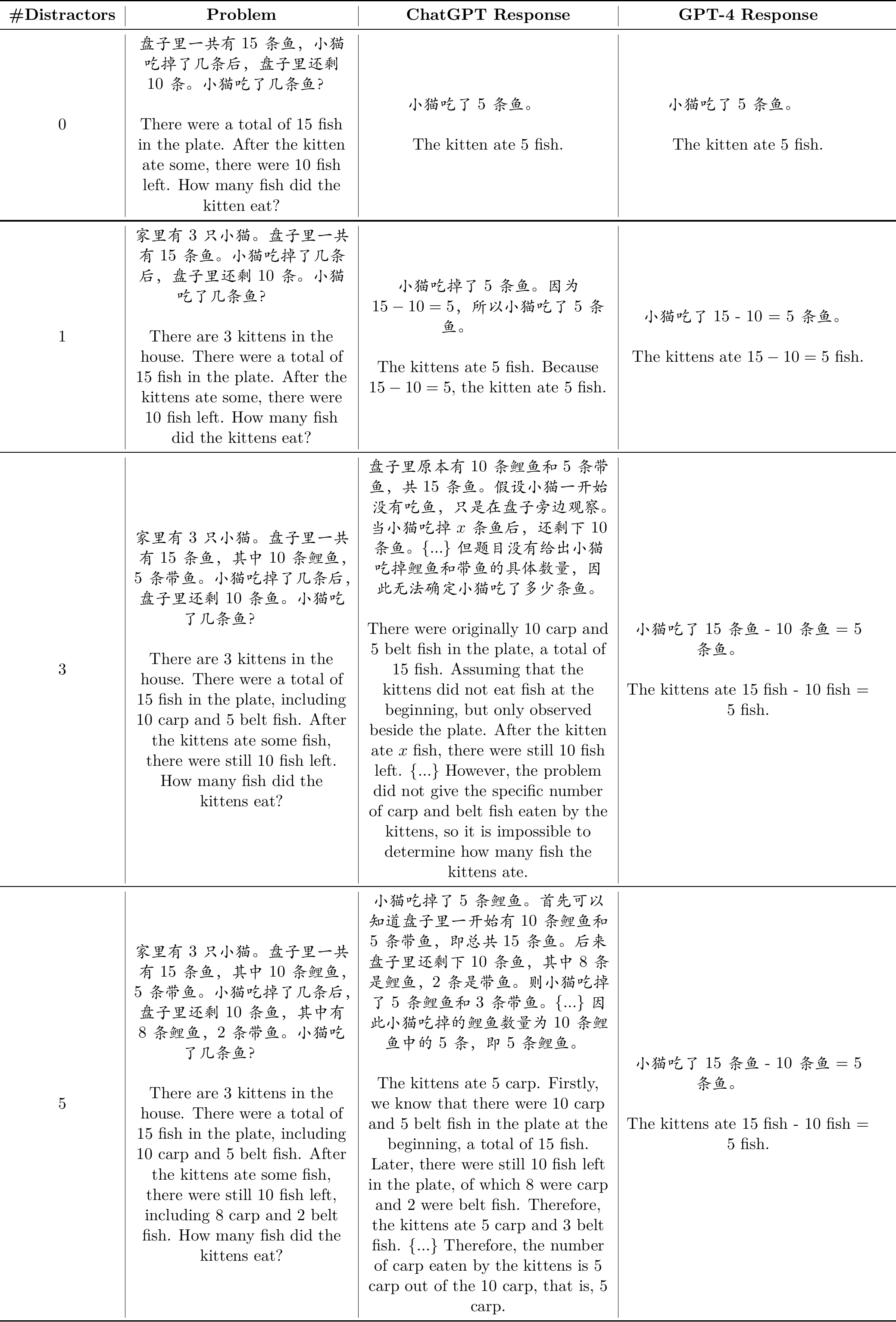}
}
\caption{An example of a math word problem augmented with distractors, alongside the corresponding responses generated by ChatGPT and GPT-4. The column labeled ``\#Distractor'' indicates the number of distractors injected into the problem. The first row displays the original problem without any distractors, while the subsequent rows show problems augmented with 1, 3, and 5 distractors, respectively. Note that the injected phrase ``There are 3 kittens in the house'' is considered as a single distractor, whereas "including 10 carp and 5 belt fish" is regarded as a combination of two distractors, as the latter contains two distracting numerals. In the table, the ChatGPT responses are cherry-picked to illustrate certain behaviors, but the GPT-4 responses are not. Upon examining the model responses, we observe that ChatGPT is sometimes influenced by the injected distractors, resulting in erroneous reasoning, while GPT-4 consistently focuses on the relevant information, thereby producing correct and concise responses.
}
\label{tab:distractor}
\end{table*}

\subsection{Arithmetic Complexity and Reasoning Complexity}
We now investigate the contributing factors for an LLM to fail in elementary level math word problems. As is introduced in Section \ref{dataset},
we focus on two quantities that are approximate measure of the arithmetic complexity and reasoning complexity of a problem, namely 
the the number of digits that an LLM needs to manipulate and the number of reasoning steps that an LLM needs to carry out in order to solve a problem.
Intuitively, problems with higher arithmetic complexity and/or reasoning complexity should be harder to solve, resulting in lower accuracy.

In Figure \ref{fig:main} (b) and (c), we plot respectively the average test accuracy against one of the complexity measure for each LLM over the entire dataset. From the figure, we observe that all models' performance declines as either of the problem complexity measures augments.  Judged from the downward slopes of the plots, it is pertinent to say that the reasoning complexity of the problem has generally a larger impact than the arithmetic complexity.  

\subsection{Robustness}
In this section, we assess the robustness of LLMs against ``irrelevant'' information, which refers to information that relates to the topic of the problem but is inconsequential or unhelpful for its resolution. This type of robustness is of particular interest because real-world problems seldom manifest in an idealized manner where all provided information is useful. Consequently, it is vital for LLMs to effectively discern the pertinent information from the problem statement and utilize it to derive a solution.

To achieve this, we manually created a small ``distractor dataset'' comprising 60 examples, 10 for each grade level. Each example consists of an original problem and five associated problems augmented with $1\sim5$ piece(s) of irrelevant information which we refer to as \emph{distractor(s)}. We require that each distractor must contain exactly one number and fit seemlessly into the original problem statement. An example is given in Table \ref{tab:distractor}. 

We tested the top-performing LLMs on the distractor dataset, and the result is plotted in Figure \ref{fig:main} (d).
From the figure, we observe that the performance of all LLMs, with the exception of GPT-4, drops drastically as the number of distractors increases. 
Notably, ChatGPT suffers an accuracy drop of 30\% for problems augmented with merely two distractors. 
In contrast, GPT-4 only experiences minor degradation.  In Table \ref{tab:distractor} we give examples of ChatGPT and GPT4 responses to the the augmented problems, revealing that the behavior of ChatGPT and GPT-4 are qualitatively different against distractors. It can be clearly seen that ChatGPT is easily distracted by the injected information, resulting in erroneous reasoning process and conclusion, while GPT-4 is able to always stick to the relevant information.

Based on this experiment, we conclude that among the models assessed in this work GPT-4 is the only one that exhibits robustness against the distractors.

\section{Related Work}
Math-related datasets are predominantly in English \citep{math, mathqa, gsm8k, mmlu}, making them unsuitable for evaluating Chinese LLMs. Among the Chinese math-related datasets, AGI-Eval \citep{agieval} and C-Eval \citep{ceval} target general-purpose, multi-disciplinary evaluation for LLMs and contain subsets specifically designed for assessing mathematical abilities. However, the math problems in these datasets, ranging from middle school to college level, are considerably more difficult than those in our CMATH dataset, rendering it challenging to accurately measure progress given the current capabilities of LLMs. Math23K \citep{math23k} and APE210K \citep{ape210k} are datasets comprising elementary school level math word problems, which are more similar to our CMATH. However, a drawback of these datasets is the absence of fine-grained annotations, such as grade, number of reasoning steps, etc., making it impossible to obtain detailed evaluation results from them.

\section{Conclusion}
This work presents CMATH, a dataset enabling fine-grained evaluation of LLMs on elementary school level math word problems. Our evaluation on CMATH
shows that all LLMs, with the exception of GPT-4, falters at a certain grade. Moreover, our investigation into the robustness of LLMs under the presence of distracting information further underscores the superior performance of GPT-4, as it remains the only model to maintain its robustness amidst such challenges. We anticipate that the this will not only expose existing limitations in LLMs' capabilities but also serve as a catalyst for their ongoing development and improvement. 

\bibliographystyle{acl_natbib}
\bibliography{nlp,llm}

\begin{thebibliography}{25}
\expandafter\ifx\csname natexlab\endcsname\relax\def\natexlab#1{#1}\fi

\bibitem[{Amini et~al.(2019)Amini, Gabriel, Lin, Koncel-Kedziorski, Choi, and
  Hajishirzi}]{mathqa}
Aida Amini, Saadia Gabriel, Peter Lin, Rik Koncel-Kedziorski, Yejin Choi, and
  Hannaneh Hajishirzi. 2019.
\newblock \href {http://arxiv.org/abs/1905.13319} {Mathqa: Towards
  interpretable math word problem solving with operation-based formalisms}.

\bibitem[{{B}aichuan inc(2023)}]{baichuan}
{B}aichuan inc. 2023.
\newblock \href {https://github.com/baichuan-inc/baichuan-7B}
  {{B}aichuan-7{B}}.
\newblock
  \url{https://github.com/baichuan-inc/baichuan-7B/blob/main/README_EN.md}.

\bibitem[{Bubeck et~al.(2023)Bubeck, Chandrasekaran, Eldan, Gehrke, Horvitz,
  Kamar, Lee, Lee, Li, Lundberg, Nori, Palangi, Ribeiro, and
  Zhang}]{gpt4_sparks}
Sébastien Bubeck, Varun Chandrasekaran, Ronen Eldan, Johannes Gehrke, Eric
  Horvitz, Ece Kamar, Peter Lee, Yin~Tat Lee, Yuanzhi Li, Scott Lundberg,
  Harsha Nori, Hamid Palangi, Marco~Tulio Ribeiro, and Yi~Zhang. 2023.
\newblock \href {http://arxiv.org/abs/2303.12712} {Sparks of artificial general
  intelligence: Early experiments with gpt-4}.

\bibitem[{Cobbe et~al.(2021)Cobbe, Kosaraju, Bavarian, Chen, Jun, Kaiser,
  Plappert, Tworek, Hilton, Nakano, Hesse, and Schulman}]{gsm8k}
Karl Cobbe, Vineet Kosaraju, Mohammad Bavarian, Mark Chen, Heewoo Jun, Lukasz
  Kaiser, Matthias Plappert, Jerry Tworek, Jacob Hilton, Reiichiro Nakano,
  Christopher Hesse, and John Schulman. 2021.
\newblock \href {http://arxiv.org/abs/2110.14168} {Training verifiers to solve
  math word problems}.

\bibitem[{Cui et~al.(2023)Cui, Yang, and Yao}]{chinese-alpaca}
Yiming Cui, Ziqing Yang, and Xin Yao. 2023.
\newblock \href {https://arxiv.org/abs/2304.08177} {Efficient and effective
  text encoding for chinese llama and alpaca}.
\newblock \emph{arXiv preprint arXiv:2304.08177}.

\bibitem[{Du et~al.(2022)Du, Qian, Liu, Ding, Qiu, Yang, and Tang}]{glm}
Zhengxiao Du, Yujie Qian, Xiao Liu, Ming Ding, Jiezhong Qiu, Zhilin Yang, and
  Jie Tang. 2022.
\newblock {GLM:} general language model pretraining with autoregressive blank
  infilling.
\newblock pages 320--335.

\bibitem[{Hendrycks et~al.(2021{\natexlab{a}})Hendrycks, Burns, Basart, Zou,
  Mazeika, Song, and Steinhardt}]{mmlu}
Dan Hendrycks, Collin Burns, Steven Basart, Andy Zou, Mantas Mazeika, Dawn
  Song, and Jacob Steinhardt. 2021{\natexlab{a}}.
\newblock \href {http://arxiv.org/abs/2009.03300} {Measuring massive multitask
  language understanding}.

\bibitem[{Hendrycks et~al.(2021{\natexlab{b}})Hendrycks, Burns, Kadavath,
  Arora, Basart, Tang, Song, and Steinhardt}]{math}
Dan Hendrycks, Collin Burns, Saurav Kadavath, Akul Arora, Steven Basart, Eric
  Tang, Dawn Song, and Jacob Steinhardt. 2021{\natexlab{b}}.
\newblock Measuring mathematical problem solving with the math dataset.

\bibitem[{Hu et~al.(2021)Hu, Shen, Wallis, Allen-Zhu, Li, Wang, Wang, and
  Chen}]{lora}
Edward~J. Hu, Yelong Shen, Phillip Wallis, Zeyuan Allen-Zhu, Yuanzhi Li, Shean
  Wang, Lu~Wang, and Weizhu Chen. 2021.
\newblock \href {http://arxiv.org/abs/2106.09685} {{L}o{RA}: Low-rank
  adaptation of large language models}.

\bibitem[{Huang et~al.(2023)Huang, Bai, Zhu, Zhang, Zhang, Su, Liu, Lv, Zhang,
  Lei, Fu, Sun, and He}]{ceval}
Yuzhen Huang, Yuzhuo Bai, Zhihao Zhu, Junlei Zhang, Jinghan Zhang, Tangjun Su,
  Junteng Liu, Chuancheng Lv, Yikai Zhang, Jiayi Lei, Yao Fu, Maosong Sun, and
  Junxian He. 2023.
\newblock C-eval: A multi-level multi-discipline chinese evaluation suite for
  foundation models.
\newblock \emph{arXiv preprint arXiv:2305.08322}.

\bibitem[{{IDEA-CCNL}(2023)}]{ziya}
{IDEA-CCNL}. 2023.
\newblock \href
  {https://huggingface.co/IDEA-CCNL/Ziya-LLaMA-13B-v1/blob/main/README.md}
  {Idea-ccnl/ziya-llama-13b-v1}.
\newblock
  \url{https://huggingface.co/IDEA-CCNL/Ziya-LLaMA-13B-v1/blob/main/README.md}.

\bibitem[{Nijkamp et~al.(2023)Nijkamp, Pang, Hayashi, Tu, Wang, Zhou, Savarese,
  and Xiong}]{codegen}
Erik Nijkamp, Bo~Pang, Hiroaki Hayashi, Lifu Tu, Huan Wang, Yingbo Zhou, Silvio
  Savarese, and Caiming Xiong. 2023.
\newblock \href {http://arxiv.org/abs/2203.13474} {Codegen: An open large
  language model for code with multi-turn program synthesis}.

\bibitem[{OpenAI(2023)}]{gpt4_report}
OpenAI. 2023.
\newblock \href {http://arxiv.org/abs/2303.08774} {{GPT}-4 technical report}.

\bibitem[{Ouyang et~al.(2022)Ouyang, Wu, Jiang, Almeida, Wainwright, Mishkin,
  Zhang, Agarwal, Slama, Ray, Schulman, Hilton, Kelton, Miller, Simens, Askell,
  Welinder, Christiano, Leike, and Lowe}]{instruct_gpt}
Long Ouyang, Jeff Wu, Xu~Jiang, Diogo Almeida, Carroll~L. Wainwright, Pamela
  Mishkin, Chong Zhang, Sandhini Agarwal, Katarina Slama, Alex Ray, John
  Schulman, Jacob Hilton, Fraser Kelton, Luke Miller, Maddie Simens, Amanda
  Askell, Peter Welinder, Paul Christiano, Jan Leike, and Ryan Lowe. 2022.
\newblock \href {http://arxiv.org/abs/2203.02155} {Training language models to
  follow instructions with human feedback}.

\bibitem[{Peng(2023)}]{rwkv-huggingface}
Bo~Peng. 2023.
\newblock \href {https://huggingface.co/BlinkDL/rwkv-4-raven} {{RWKV}-4-raven}.
\newblock \url{https://huggingface.co/BlinkDL/rwkv-4-raven}.

\bibitem[{Peng et~al.(2023)Peng, Alcaide, Anthony, Albalak, Arcadinho, Cao,
  Cheng, Chung, Grella, GV, He, Hou, Kazienko, Kocon, Kong, Koptyra, Lau,
  Mantri, Mom, Saito, Tang, Wang, Wind, Wozniak, Zhang, Zhang, Zhao, Zhou, Zhu,
  and Zhu}]{RWKV}
Bo~Peng, Eric Alcaide, Quentin Anthony, Alon Albalak, Samuel Arcadinho, Huanqi
  Cao, Xin Cheng, Michael Chung, Matteo Grella, Kranthi~Kiran GV, Xuzheng He,
  Haowen Hou, Przemyslaw Kazienko, Jan Kocon, Jiaming Kong, Bartlomiej Koptyra,
  Hayden Lau, Krishna Sri~Ipsit Mantri, Ferdinand Mom, Atsushi Saito, Xiangru
  Tang, Bolun Wang, Johan~S. Wind, Stansilaw Wozniak, Ruichong Zhang, Zhenyuan
  Zhang, Qihang Zhao, Peng Zhou, Jian Zhu, and Rui-Jie Zhu. 2023.
\newblock \href {http://arxiv.org/abs/2305.13048} {{RWKV}: Reinventing {RNN}s
  for the transformer era}.

\bibitem[{Sun and Qiu(2023)}]{moss}
Tianxiang Sun and Xipeng Qiu. 2023.
\newblock \href {https://github.com/OpenLMLab/MOSS/blob/main/README_en.md}
  {{MOSS}}.
\newblock {G}ithub.
\newblock \url{https://github.com/OpenLMLab/MOSS/blob/main/README_en.md}.

\bibitem[{Taylor et~al.(2022)Taylor, Kardas, Cucurull, Scialom, Hartshorn,
  Saravia, Poulton, Kerkez, and Stojnic}]{galactica}
Ross Taylor, Marcin Kardas, Guillem Cucurull, Thomas Scialom, Anthony
  Hartshorn, Elvis Saravia, Andrew Poulton, Viktor Kerkez, and Robert Stojnic.
  2022.
\newblock \href {http://arxiv.org/abs/2211.09085} {Galactica: A large language
  model for science}.

\bibitem[{{THUDM}(2023{\natexlab{a}})}]{chatglm6b}
{THUDM}. 2023{\natexlab{a}}.
\newblock \href {https://github.com/THUDM/ChatGLM-6B/blob/main/README_en.md}
  {{C}hat{GLM}-6{B}}.
\newblock \url{https://github.com/THUDM/ChatGLM-6B/blob/main/README_en.md}.

\bibitem[{{THUDM}(2023{\natexlab{b}})}]{chatglm6b-v2}
{THUDM}. 2023{\natexlab{b}}.
\newblock \href {https://github.com/THUDM/ChatGLM2-6B/blob/main/README_EN.md}
  {{C}hat{GLM}2-6{B}}.
\newblock \url{https://github.com/THUDM/ChatGLM2-6B/blob/main/README_EN.md}.

\bibitem[{Wang et~al.(2017)Wang, Liu, and Shi}]{math23k}
Yan Wang, Xiaojiang Liu, and Shuming Shi. 2017.
\newblock \href {https://doi.org/10.18653/v1/D17-1088} {Deep neural solver for
  math word problems}.
\newblock In \emph{Proceedings of the 2017 Conference on Empirical Methods in
  Natural Language Processing}, pages 845--854, Copenhagen, Denmark.
  Association for Computational Linguistics.

\bibitem[{Wei et~al.(2023)Wei, Wang, Schuurmans, Bosma, Ichter, Xia, Chi, Le,
  and Zhou}]{cot}
Jason Wei, Xuezhi Wang, Dale Schuurmans, Maarten Bosma, Brian Ichter, Fei Xia,
  Ed~Chi, Quoc Le, and Denny Zhou. 2023.
\newblock \href {http://arxiv.org/abs/2201.11903} {Chain-of-thought prompting
  elicits reasoning in large language models}.

\bibitem[{Zeng et~al.(2022)Zeng, Liu, Du, Wang, Lai, Ding, Yang, Xu, Zheng, Xia
  et~al.}]{glm130}
Aohan Zeng, Xiao Liu, Zhengxiao Du, Zihan Wang, Hanyu Lai, Ming Ding, Zhuoyi
  Yang, Yifan Xu, Wendi Zheng, Xiao Xia, et~al. 2022.
\newblock {GLM}-130b: An open bilingual pre-trained model.
\newblock \emph{arXiv preprint arXiv:2210.02414}.

\bibitem[{Zhao et~al.(2020)Zhao, Shang, Liu, Wang, and Liu}]{ape210k}
Wei Zhao, Mingyue Shang, Yang Liu, Liang Wang, and Jingming Liu. 2020.
\newblock \href {http://arxiv.org/abs/2009.11506} {{Ape}210k: A large-scale and
  template-rich dataset of math word problems}.

\bibitem[{Zhong et~al.(2023)Zhong, Cui, Guo, Liang, Lu, Wang, Saied, Chen, and
  Duan}]{agieval}
Wanjun Zhong, Ruixiang Cui, Yiduo Guo, Yaobo Liang, Shuai Lu, Yanlin Wang, Amin
  Saied, Weizhu Chen, and Nan Duan. 2023.
\newblock \href {http://arxiv.org/abs/2304.06364} {Agieval: A human-centric
  benchmark for evaluating foundation models}.

\end{thebibliography}
\appendix

\end{document}